\documentclass[sigconf, nonacm]{acmart}
\AtBeginDocument{%
  }

\setcopyright{none}
\acmConference[Copyright by author(s)]{}{2025}{}
\acmISBN{}

\usepackage{graphicx}
\usepackage{booktabs} 
\usepackage{hyperref}

\usepackage{amsmath}
\usepackage{amsthm}
\usepackage{algorithm}
\usepackage{algorithmic}
\usepackage{mathtools}
\usepackage[capitalize,noabbrev]{cleveref}
\usepackage{graphicx}
\usepackage{booktabs}
\usepackage{wrapfig}
\usepackage{lipsum}
\usepackage{booktabs}
\usepackage{xspace}
\usepackage{enumitem}
\usepackage{hyperref}
\usepackage{caption}
\usepackage{pifont}
\usepackage{subcaption}
\usepackage{subcaption}
\usepackage{orcidlink}
\usepackage{xcolor,colortbl}
\usepackage{natbib}
\usepackage{enumitem}
\newcommand{\name}{HyperCore\xspace}
\usepackage{orcidlink}
\newcommand{\x}{\mathbf{x}}
\newcommand{\y}{\mathbf{y}}
\newcommand{\z}{\mathbf{z}}
\newcommand{\R}{\mathbb{R}}
\usepackage[frozencache,cachedir=.]{minted} 
\usepackage{xcolor} 
\newcommand{\cmark}{\ding{51}}
\newcommand{\xmark}{\ding{55}}
\definecolor{bg}{rgb}{0.95,0.95,0.95} 

\usepackage[english]{babel}
\usepackage{hyperref}       
\renewenvironment{proof}{\begin{newproof}}{\end{newproof}\qed}

\theoremstyle{plain}

\theoremstyle{definition}

\theoremstyle{remark}

\begin{document}

\title{\name: The Core Framework for Building Hyperbolic Foundation Models with Comprehensive Modules} 

\author{Neil He}
\email{neilhe6345@gmail.com}
\orcid{0009-0008-3193-2448}
\affiliation{%
  \institution{Yale University}
  \city{New Haven}
  \country{United States}
}
\author{Menglin Yang}
\email{mlyang.yale@outlook.com}
\orcid{0000-0003-2510-5282}
\affiliation{%
  \institution{Yale University}
  \city{New Haven}
  \country{United States}
}
\author{Rex Ying}
\email{rex.ying@yale.edu}
\orcid{0000-0002-5856-5229}
\affiliation{%
  \institution{Yale University}
  \city{New Haven}
  \country{United States}
}

\renewcommand{\shortauthors}{Neil He et al.}

\begin{abstract}
  Hyperbolic neural networks have emerged as a powerful tool for modeling hierarchical data across diverse modalities. Recent studies show that token distributions in foundation models exhibit scale-free properties, suggesting that hyperbolic space is a more suitable ambient space than Euclidean space for many pre-training and downstream tasks. 
  However, existing tools lack essential components for building hyperbolic foundation models, making it difficult to leverage recent advancements. 
  We introduce \textit{\name}, a comprehensive open-source framework that provides core modules for constructing hyperbolic foundation models across multiple modalities.
  \name’s modules can be effortlessly combined to develop novel hyperbolic foundation models, eliminating the need to extensively modify Euclidean modules from scratch and possible redundant research efforts.
  To demonstrate its versatility, we build and test the first fully hyperbolic vision transformers (LViT) with a fine-tuning pipeline, the first fully hyperbolic multi-modal CLIP model (L-CLIP), and a hybrid Graph RAG with a hyperbolic graph encoder.
  Our experiments demonstrate that LViT outperforms its Euclidean counterpart. Additionally, we benchmark and reproduce experiments across hyperbolic GNNs, CNNs, Transformers, and vision Transformers to highlight \name’s advantages.\footnote{Code is available at \href{https://github.com/Graph-and-Geometric-Learning/HyperCore}{\textcolor{blue}{https://github.com/Graph-and-Geometric-Learning/HyperCore}}}
\end{abstract}



\maketitle

\section{Introduction}
\label{sec:intro}
Recent efforts have seen rapid advances in developing neural network architectures for non-Euclidean spaces~\cite{peng2021hyperbolic,mettes2023hyperbolic,yang2022hyperbolicsurvey}, especially for hyperbolic neural networks~\cite{HNN,HNN++,van2023poincar,hgcn2019,liu2019HGNN,Bdeir2024fully,chen2021fully}.  As their negative curvature enables for exponential growth in volume with respect to distance, hyperbolic spaces are particularly suitable for embedding and representing hierarchical data, while their Euclidean counterparts suffer from embedding distortions~\cite{2010hyperbolic,sarkar2011low}. As a result, many recent works have shown that embedding data in hyperbolic manifolds have the potential to improve expressiveness and efficiency in representation learning for a wide range of applications, with significant advancements in domains such as languages~\cite{nickel2017poincare,nickel2018learning,dhingra2018embedding,chen2021probing}, graphs~\cite{hgcn2019,liu2019HGNN,zhang2021hyperbolic,yang2023hyperbolic,yang2023kappahgcn,yang2024hypformer}, social networks~\cite{yang2021discrete,yang2022htgn}, recommendation systems~\cite{yang2022hrcf,yang2022hicf,chen2021modeling,sun2021hgcf}, image embedding~\cite{khrulkov2020hyperbolic,guo2022clipped,ermolov2022hyperbolic}, and segmentation~\cite{atigh2022hyperbolic,weng2021unsupervised,chen2023hyperbolic}.

In large-scale unsupervised learning such as casual language models, as token distribution has been shown to exhibit scale free properties~\cite{coenen2019visualizing, yang2024hyplora}, hyperbolic foundation models are especially desirable as they have been shown to be able to distinguish low-level tokens using low-dimension embeddings~\cite{HNN++, yang2024hyplora, chen2021fully}. Recent studies have made significant advances to the development of hyperbolic foundation models, including Transformers~\cite{gulcehre2019hyperbolicAT, HNN++, chen2021fully} with hyperbolic attention mechanisms, hyperbolic Transformers with linear attention~\cite{yang2024hypformer}, hyperbolic vision Transformers~\cite{fein-ashley2024hvt}, hyperbolic pre-trained large language models ~\cite{chen2024hyperbolic}, hyperbolic fine-tuning methods for LLMs~\cite{yang2024hyplora}, and hyperbolic vision-language models~\cite{desai2023hyperbolic, pal2025compositional}.

\textbf{Limitations of Existing Tools.} Despite these efforts, there is still no comprehensive open-source framework that enables one to easily utilize recently developed modules and methods for building hyperbolic foundation models. Implementing hyperbolic models is non-trivial, as it requires extensive expertise in differential geometry and careful attention to ensure model stability. Existing implementations are often custom-tailored for specific use cases and are incompatible with one another.
Libraries that have been developed for optimization and computation on manifolds~\cite{geomstats, kochurov2020geoopt, manifolds.jl} do not implement the common hyperbolic neural networks layers. While some hyperbolic deep learning frameworks do exist~\cite{spengler2023hypll, hyperlib}, their scope is limited in building hyperbolic foundation models: \begin{enumerate}[wide,label=(\roman*)]
    \item \textbf{Limited Modules.} Existing tools such as HypLL~\cite{spengler2023hypll} and Hyperlib~\cite{hyperlib} lack many essential components for building hyperbolic foundation models, including but not limited to hyperbolic Transformers, vision Transformers, fine-tuning of hyperbolic foundation models, recent GNNs and graph Transformers, and more.
    \item \textbf{Inflexibility and Unintuitive-Usage.} Prior libraries like Hyperlib defined many hyperbolic operations through manifold operations instead of accessible modules, making it difficult to build a hyperbolic neural network without an abundance of knowledge in hyperbolic geometry. Even though HypLL emphasizes its usability, the fact that only basic neural network operations are implemented results in the tool being \textit{inflexible} --- significant modifications to the modules are necessary for constructing models beyond variants of simpler models such as linear networks and CNNs.
    \item \textbf{Limited Model Support.} HypLL only supports hyperbolic operations on the Poincar{\'e} ball, leaving out support for hyperbolic foundation models developed for the Lorentz hyperboloid, such as hyperbolic BERT~\cite{chen2021fully} and linear attention~\cite{yang2024hypformer}.
\end{enumerate}
\textbf{Challenges.} 
Designing a framework for hyperbolic foundation models that is comprehensive and simple to use goes beyond merely collecting previously developed methods. We overcame key engineering challenges:
\begin{enumerate}
    \item [(1)] \textbf{Modules with Variable Curvatures.} While some previous methods proposed per-layer variable curvatures in different architectures~\cite{yang2024hypformer, hgcn2019, lgcn}, these approaches are not integrated in many modules for foundation models, such as convolutional layers and batch normalizations~\cite{van2023poincar, Bdeir2024fully}. Many works also fix the curvature to $-1$ in practice~\cite{chen2021fully, dai2021H2H, zhu2020gil}. Our framework have integrated support for variable curvature learning whenever possible, taking full advantage of the representation flexibility in hyperbolic spaces. Existing frameworks do not have such support: neither Hyperlib nor HypLL have integrated per-layer-change-of-curvature support in their neural network layers.
    \item [(2)]\textbf{Adapting Methods to Modules.} While the building block operations have already been developed, extending these methods to fully hyperbolic modules require further efforts to modify and improve upon them. For instance, the hybrid fine-tuning in HypLoRA needs to be extended to a fully hyperbolic case by incorporating hyperbolic operation not only in low-rank matrix multiplications but also in feature combination that follows. Additional examples are extending convolution to patch embeddings and finding the inverse of hyperbolic concatenation~\cite{qu2022hyperbolic} for truncation.
\end{enumerate}  
\begin{table*}[]
\begin{tabular}{@{}lcccccccccc@{}}
\toprule
\textbf{Framework}      & \textbf{MLPs}          & \textbf{GNNs} & \textbf{CNNs} & \textbf{Transformers} & \textbf{ViTs} & \textbf{Fine Tuning}& \textbf{CLIP} & \textbf{Graph RAG }&\textbf{$\mathbb{L}^{n, K}$}& \textbf{$\mathbb{P}^{n, K}$}  \\ \midrule
HypLL~\cite{spengler2023hypll}          &  \cmark     & \xmark     &  \cmark                       &     \xmark                         &\xmark        &\xmark &\xmark &\xmark &\xmark        &  \cmark        \\
Hyperlib~\cite{hyperlib}     &  \cmark       &  \cmark            &   \xmark                          &      \xmark                       &     \xmark &\xmark &\xmark &\xmark & \cmark &  \cmark                  \\
\textbf{\name} &  \cmark        &  \cmark               &   \cmark                   &           \cmark                  &     \cmark &\cmark &\cmark&\cmark    &\cmark &\cmark            \\ \bottomrule
\end{tabular}\caption{Models that can be built using the modules supported by \name and other existing hyperbolic deep learning frameworks. $\mathbb{L}^{n, K}$ and $\mathbb{P}^{n, K}$ indicate the Lorentz space and Poincar{\'e} Ball model of dimension $n$ and curvature $K$ respectively. \name can be used to implement a far more extensive list of hyperbolic foundation models.}\label{comparison_table}
\end{table*}
\textbf{\name.} We introduce \textit{\name}, a core hyperbolic deep learning framework equipped with a comprehensive list of accessible modules designed for building foundation models in hyperbolic space, built on top of PyTorch for familiar and intuitive usage. Compared to existing frameworks, \name has several key advantages, particularly in its completeness and scope: 
\begin{enumerate}[wide,label=(\roman*)]
    \item \textbf{Flexible and Intuitive Foundation Model Support.} \name supports intuitive construction of pre-training and fine-tuning pipelines of hyperbolic foundation models. It is capable of doing much more than reproducing existing  models---its components can be effortlessly combined to construct novel hyperbolic models foundation models that have yet to be proposed. We construct several new hyperbolic foundation models in \cref{hvit}, demonstrating the power of \name in simplifying future works. 
    \item \textbf{Comprehensive Modules and Model Support.} \name provides a comprehensive list of essential hyperbolic modules for building a wide range of hyperbolic foundation models for learning across diverse modalities. 
    \name also supports both the Lorentz hyperbolic space and Poincar{\'e} ball model, not being limited to one or the other. 
    In \cref{comparison_table}, we show a comparison between the supported hyperbolic models users can build using \name and existing frameworks. \name can be used to build a far more extensive list of hyperbolic foundation models.  
    \item \textbf{User Accessibility}. \name is easy to use not only for hyperbolic deep learning researchers but also the more general AI audience. As the API is designed almost identically to Euclidean counterparts, users only need a high-level understanding of foundation models to build hyperbolic ones using \name.
\end{enumerate} 
We perform extensive empirical studies to demonstrate \name's advantages: (1) To demonstrate \name's \textit{flexibility and intuitive foundation model support}, we build and test the first \textit{fully hyperbolic fine-tuning pipeline for vision transformers (LViT)} in Lorentz space, demonstrating that it outperforms the Euclidean ViT in image classification. We also build the first \textit{hyperbolic CLIP model (L-CLIP)} and evaluate its performance on image-text retrieval tasks, demonstrating \name's capability to build multi-modal hyperbolic foundation models. We also extend the Euclidean Graph RAG model~\cite{he2024gretriever} to incorporate a hyperbolic GNN with hyperbolic fine-tuning (HypGraphRAG). Details can be found in \cref{hvit}. (2) To demonstrate \name's \textit{comprehensiveness}, we reproduce and benchmark an extensive list of prior studies across CNNs, GNNs, Transformers, and vision Transformers using \name.

\textbf{Future: Going Beyond Finding Hyperbolic Counterparts.} A large body of works in the hyperbolic deep learning communities focus on developing the hyperbolic analogue of popular Euclidean models, such as GNNs~\cite{hgcn2019, liu2019HGNN}, CNNs~\cite{Bdeir2024fully, van2023poincar}, and Transformers~\cite{gulcehre2019hyperbolicAT, fein-ashley2024hvt}. As demonstrated by our constructions of LViT and L-CLIP, \name enables effortless development of new hyperbolic foundation models, significantly reducing the effort required to design hyperbolic counterparts of Euclidean architectures. As a result, \name allows future studies to instead focus on deeper analyses of hyperbolic foundation models. Here we suggest a few directions for future analysis: 
\begin{enumerate}
    \item [(1)] \textbf{Curvature Sensitivity. } Curvature is the defining characteristic of hyperbolic space, yet its impact on model behavior remains underexplored. Many studies have either fixed the curvature (often at $-1$)~\cite{zhu2020gil, chen2021fully, dai2021H2H} or set it as a learnable parameter~\cite{desai2023hyperbolic}, offering little insight into how curvature affects model performance. Greater understanding of this essential geometric attribute could significantly improve future hyperbolic foundation models.
    \item [(2)] \textbf{Geometric Understanding of Modules. } Many hyperbolic modules are designed as projections of their Euclidean counterparts. For example, in Lorentz space, models often perform Euclidean operations on the time-like dimension and project the output back onto the manifold by computing its space-like component~\cite{chen2021fully, Bdeir2024fully, yang2024hypformer}. While in the case of linear transformations this approach covers both Lorentz boost and Lorentz rotation~\cite{chen2021fully}, the geometric meaning of non-linear operations implemented in this manner are usually not discussed. Better understanding of the geometric implications of these methods and their effects on model behavior could lead to more principled and effective designs.
    \item [(3)] \textbf{Hyperbolic Training Schemes.} The training dynamics of hyperbolic foundation models remain poorly understood, particularly in the context of pre-training large-scale models with billions or trillions of parameters. Hyperbolic models may benefit from specialized training strategies tailored to their unique geometry, rather than relying on conventional Euclidean approaches.
\end{enumerate}  

\textbf{Contribution.} The specifics of this study can be summarized as follows: (1) We introduce \textit{\name}, a core framework for hyperbolic foundation models, equipped with comprehensive modules that streamline hyperbolic foundation model construction. By overcoming key engineering challenges, \name surpasses existing tools in its scope and functionality, enabling effortless development of new hyperbolic foundation models. This not only simplifies future research but also minimizes potential redundant research efforts; (2) We develop and evaluate the fully first hyperbolic Vision Transformer (LViT) with a fine-tuning pipeline, the first fully hyperbolic CLIP model (L-CLIP), and the first hyperbolic GraphRAG model, demonstrating \name's versatility in constructing new hyperbolic foundation models and its capability to support multi-modal architectures; (3) To showcase \name's comprehensiveness, we benchmark and reproduce diverse prior works across graph, image, and language modalities. This includes extensive evaluations of hyperbolic GNNs on standard link prediction, node classification, and graph reconstruction tasks—the latter rarely explored in previous studies. They highlight \name's capability to generate novel insights into existing methods.

\section{Related Works}
\textbf{Hyperbolic Modules and Foundation Models.} Recent studies have made significant progress toward building hyperbolic foundation models. HNN~\cite{HNN} and HNN++\cite{HNN++} developed many basic operations, such as hyperbolic linear and convolutional layers, and multinomial logistic regression (MLR). HGCN~\cite{hgcn2019} and HGNN~\cite{liu2019HGNN} were then among the first to adapt graph neural networks (GNNs) to hyperbolic spaces. 
More recently, HyboNet~\cite{chen2021fully} proposed a framework of hyperbolic neural networks that does not depend on the Euclidean tangent spaces; Poincar{\'e} ResNet~\cite{van2023poincar} and HCNN~\cite{Bdeir2024fully} developed essential components for hyperbolic computer vision models; Hypformer~\cite{yang2024hypformer} developed numerous essential hyperbolic modules to propose an efficient hyperbolic Transformer model that operates directly on the manifold; LResNet~\cite{he2025lresnet} proposed an efficient and stable Lorentzian residual connection method, and HypLoRA extended the LoRA fine-tuning methods to Lorentz space~\cite{yang2024hyplora}. Several works have also developed essential module such as hyperbolic entailment cone and contrastive loss for contrastive learning~\cite{desai2023hyperbolic, pal2025compositional}.

\textbf{Hyperbolic Deep Learning Tools.}
A few frameworks have been built to simply deep learning in hyperbolic space. Geoopt~\cite{kochurov2020geoopt} and Geomstats~\cite{geomstats} focus on implement manifold operations and manifold optimization techniques. For building hyperbolic neural networks, HypLL~\cite{spengler2023hypll} is a hyperbolic deep learning library focused on ease of use and debugging. Hyperlib~\cite{hyperlib} is another library that focuses more on implementing hyperbolic layers and embedding methods. However, these tools have limitations as discussed earlier. \cref{comparison_table} shows a comparison between the hyperbolic supported by \name and these existing hyperbolic deep learning frameworks, where we can see that \name can be used to build a far more extensive list of hyperbolic foundation models.

\section{Framework Overview}
\begin{figure*} 
    \centering
        \centering
       \includegraphics[width=\linewidth]{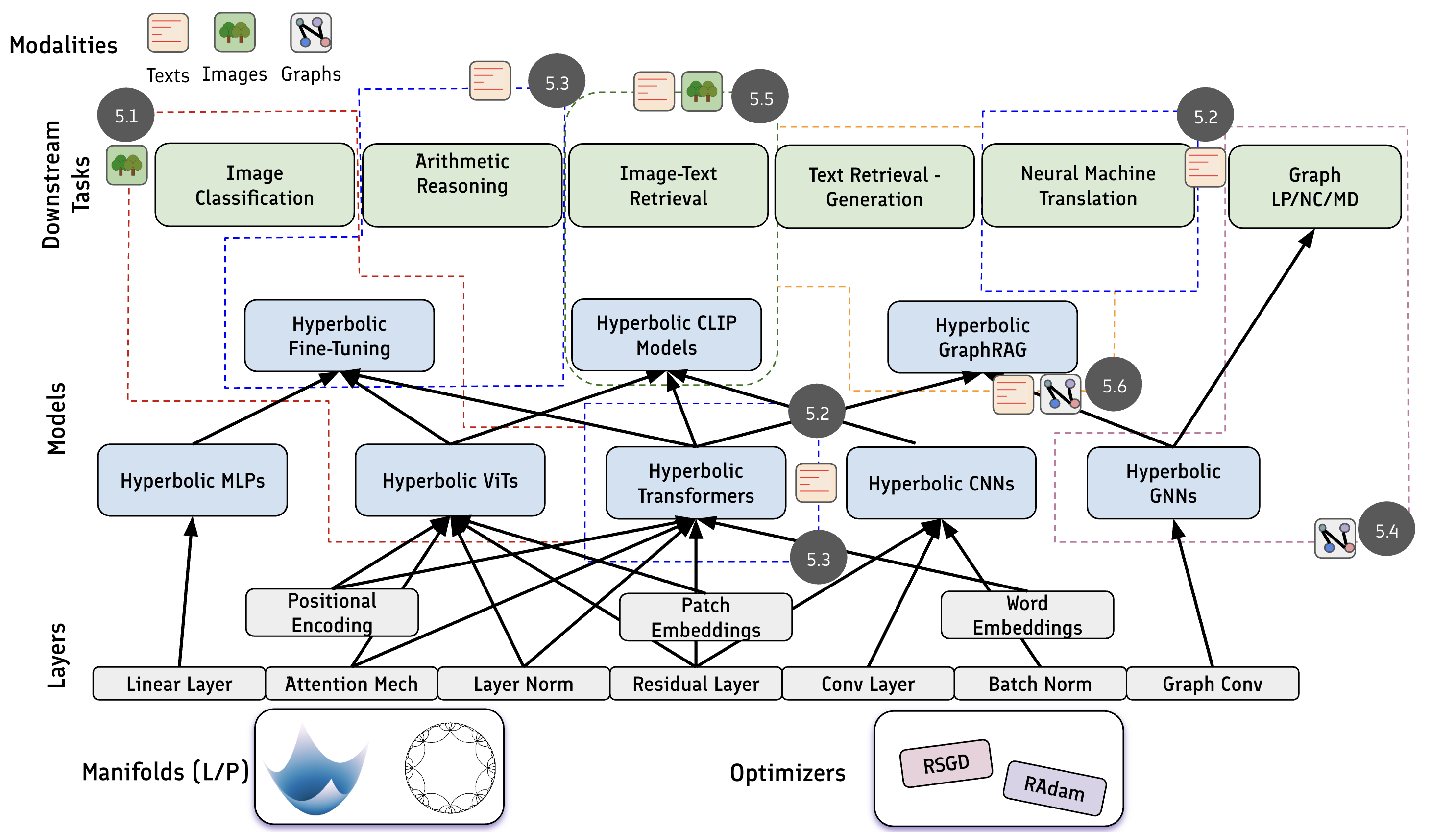}
       \caption{Snapshot of high level categories of hyperbolic foundation models supported by \name and essential modules in \name used by these models, along with example downstream tasks and the data modalities that interact with these models. Boxes drawn with dashed lines encloses a downstream task with the its associated model. For graph tasks in 5.4, LP, NC, and MD stand for link prediction, node classification, and minimizing distortion in graph reconstruction tasks respectively. The relevant subsections in \cref{eval} for each model-task combination are also indicated in white.}
       \label{model_graph}3
\end{figure*}

In this section, we give an overview of \name's framework and the supported modules. \name provides numerous functionalities, from manifold supports to hyperbolic neural network layers and optimizers. \cref{model_graph} shows a demonstration of a snapshot \name's overall framework, with high level categories of hyperbolic foundation models supported by \name and the layers used by these models, along with some examples of downstream tasks for the models and the associated modality. \cref{model_graph} is not an exhaustive list. Instead, it serves as a snapshot of the modules we will discuss in this section and the models that will appear later. These components work together to allow for the construction of new hyperbolic foundation models in \cref{hvit}.

\subsection{The \texttt{nn} module}
The heart of \name is the neural network layers in the \texttt{nn} module. These modules are designed by overcoming some of the challenges mentioned in \cref{sec:intro}.

\textbf{Linear Layers.} We support several basic hyperbolic linear layers that are foundational to building hyperbolic neural networks~\cite{HNN, HNN++, chen2021fully, yang2024hypformer}. Additionally, we provide tangent space-based activation methods~\cite{HNN} as well as activation methods that operate directly on the Lorentz hyperboloid~\cite{yang2024hypformer}. 

For classification tasks, we also provide the parameter-reduced Poincar{\'e} multinomial logistic regression (MLR) layer~\cite{HNN++} and the Lorentzian MLR layer~\cite{Bdeir2024fully}.

\textbf{Coordinate Splitting and Concatenation.} Additional fundamental hyperbolic operations for building hyperbolic neural networks are the splitting and concatenation of hyperbolic coordinates. \name supports the splitting and concatenation methods in Poincar{\'e} ball model and Lorentz~\cite{van2023poincar, qu2022autoencoding}. 

\textbf{Convolutional and Residual Networks.} We provide support to building hyperbolic CNNs and ResNets in both the Poincar{\'e} ball model and the Lorentz hyperboloid model. To this end, we provide the implementation for Poincar{\'e} and Lorentz convolutional layers~\cite{HNN++, van2023poincar, Bdeir2024fully}. The latter was modified to use the more general linear layer to allow for varying curvatures per layer~\cite{yang2024hypformer}.

For building neural networks with residual connections, we provide an implementation for the parallel transport method for residual connection available for both manifolds~\cite{van2023poincar, Katsman2023RiemannianRN}. We also provide the more efficient and numerically stable residual connection method for Lorentz hyperboloid recently proposed in LResNet~\cite{he2025lresnet}.

\textbf{Batch and Layer Normalization, and Pooling.} We also provide the support for Poincar{\'e} and Lorentzian batch normalization layers~\cite{van2023poincar, Bdeir2024fully}, as well as the layer normalization layer~\cite{yang2024hypformer}. Additionally, we provide a fully hyperbolic global pooling layer for building CNNs in Lorentz space~\cite{Bdeir2024fully} and a tangent space pooling layer for both manifolds.

\textbf{Attention Mechanisms and Transformer modules.} Several recent advancements have been made toward developing hyperbolic attention mechanisms for building hyperbolic Transformers, which are also missing in existing libraries such as HypLL. \name provides support for these operations, including the several softmax self-attention mechanism~\cite{gulcehre2019hyperbolicAT, HNN++, chen2021fully} and Lorentz linear attention mechanism ~\cite{yang2024hypformer}. 

Additionally, we also provide hyperbolic positional encoding layers~\cite{yang2024hypformer, fein-ashley2024hvt}, hyperbolic word embedding, as well as patch embedding implemented through hyperbolic convolutional layers We also implement hyperbolic fine-tuning methods using  hypLoRA~\cite{yang2024hyplora}.
 
\textbf{Graph and Neighborhood Aggregation.}
Extensive research has been done to develop hyperbolic learning frameworks for learning on tree-like graphs, achieving significant improvements over their Euclidean counterparts. In support of building hyperbolic graph foundation models, \name supports a comprehensive list of hyperbolic graph encoding layers and neighborhood aggregation methods proposed in previous studies, which is missing in existing libraries such as HypLL. These include HGCN~\cite{hgcn2019}, HGNN~\cite{liu2019HGNN}, LGCN~\cite{lgcn}, HGAT~\cite{zhang2021hyperbolic}, H2HGCN~\cite{dai2021H2H}, HyboNet~\cite{chen2021fully}, and GIL~\cite{zhu2020gil}. 

\begin{listing}[ht]
\caption{Example of building a fully hyperbolic Transformer encoder/decoder block.}\label{example}
\begin{minted}[frame=lines,framesep=1mm,baselinestretch=1,fontsize=\small,bgcolor=bg]{python}
import torch
import torch.nn as nn
import hypercore.nn as hnn

class LTransformerBlock(nn.Module):
    def __init__(self, manifold, d_model: int, n_head: int):
        super().__init__()
        dim_per_head = d_model // n_head
        out_dim = d_model - 1
        mlp_dim = d_model * 4 - 1
        self.manifold = manifold
        self.attn = hnn.LorentzMultiheadAttention(manifold, 
            dim_per_head, dim_per_head, n_head, 
            attention_type='full', trans_heads_concat=True)
        self.ln_1 = hnn.LorentzLayerNorm(manifold, out_dim)
        self.mlp = nn.Sequential(
            hnn.LorentzLinear(manifold, d_model, mlp_dim),
            hnn.LorentzActivation(manifold, 
                activation=nn.GELU()),
            hnn.LorentzLinear(manifold, mlp_dim+1, out_dim),
        )
        self.ln_2 = hnn.LorentzLayerNorm(manifold, out_dim)
        self.res1 = hnn.LResNet(manifold, use_scale=True)
        self.res2 = hnn.LResNet(manifold, use_scale=True)

    def forward(self, x, attn_mask=None):
        lx = self.ln_1(x)
        ax = self.attn(lx, lx, mask=attn_mask)
        x = self.res1(x, ax)
        x = self.res2(x, self.mlp(self.ln_2(x)))
        return x
\end{minted}
\end{listing}
\subsection{The \texttt{manifolds} and \texttt{optimizer} Modules}
\name provides support for handling hyperbolic manifold operations and optimization of hyperbolic parameters. Within the framework, these operations enable the modules in the previous section. We detail the relevant components here.

\textbf{Optimizers.} We provide Riemannian optimizers~\cite{becigneul2018riemannian} for training hyperbolic parameters, built on top of Geoopt~\cite{kochurov2020geoopt} to utilize its support and optimizations for manifold tensors. Additionally, many previous employed training strategies to train Euclidean and manifold parameters using different attributes~\cite{chen2021fully, Bdeir2024fully, yang2024hypformer}. To simply this procedure, we provide support to automatically train of Euclidean and hyperbolic parameters on separate training schemes, allowing for more training flexibility.  

\textbf{Manifolds.} \name currently supports both the Lorentzian hyperboloid and Poincar{\'e} ball formulations of hyperbolic space, with implementations provided within the \texttt{manifolds} module. The manifold implementation is built on top of the existing, well-tested implementations in Geoopt~\cite{kochurov2020geoopt}. On top of supporting basic hyperbolic computations, the manifolds were expended to incorporate advanced operations such as weighted Poincar{\'e} and Lorentzian centroids, hyperbolic entailment cones, and coordinate manipulations that were often used in recent hyperbolic models~\cite{HNN++, chen2021fully, desai2023hyperbolic, Bdeir2024fully}. 

\subsection{Example Usage and Pre-Built Models}
\name is built to resemble PyTorch in usage to ensure seamless developer experiences. In \cref{example}, we demonstrate how to create a fully hyperbolic Transformer encoder/decoder block. Every operation here is done entirely in hyperbolic space without mapping to the tangent space; the output dimension of each layer is the \textit{manifold dimension} since the Lorentz hyperboloid is an $n$-dimensional manifold embedded in an $n+1$-dimensional ambient space. The \texttt{manifold} parameter specify the manifold (and its curvature) the network lives in, which allows \name to handle the hyperbolic operations behind the layers. For users who wish to simply try using the latest hyperbolic deep learning models, we provide pre-built models ready to be used, similar in spirit to PyTorch. We plan on adding pre-trained models on Hunggingface to make using hyperbolic models as simple as possible.

\begin{figure*} 
    \centering
        \centering
       \includegraphics[width=0.80\linewidth]{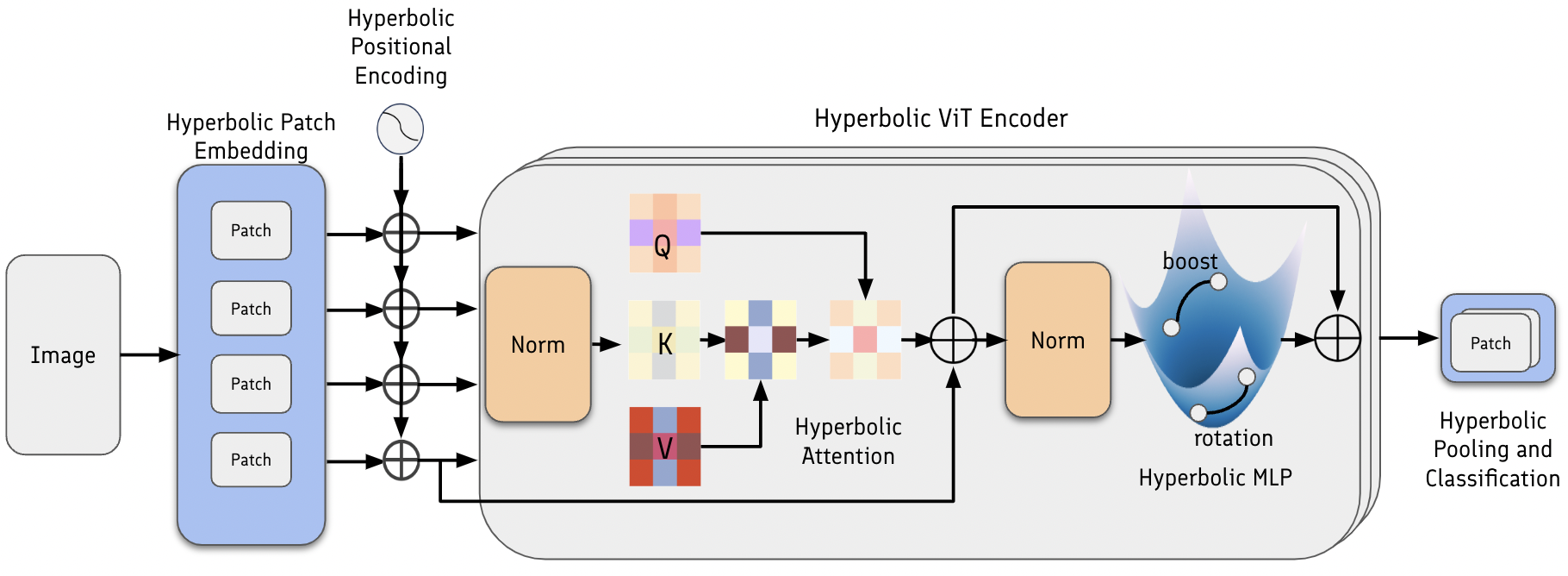}
       \caption{Framework of hyperbolic vision Transformer in Lorentz space (LViT). Images are projected into the Lorentz space and then process though hyperbolic patch embedding via a Lorentz convolutional layer. The patch embeddings are then combined with learned hyperbolic positional embeddings, where the results are then passed through a hyperbolic ViT encoder consisting of hyperbolic self-attention mechanisms and hyperbolic MLPs. Finally, the encoder outputs are averaged across patches and passed to a classifier. Dropout and activation are omitted for brevity.}
       \label{fig:HViT}
\end{figure*}

\begin{figure} 
    \centering
        \centering
       \includegraphics[width=0.85\linewidth]{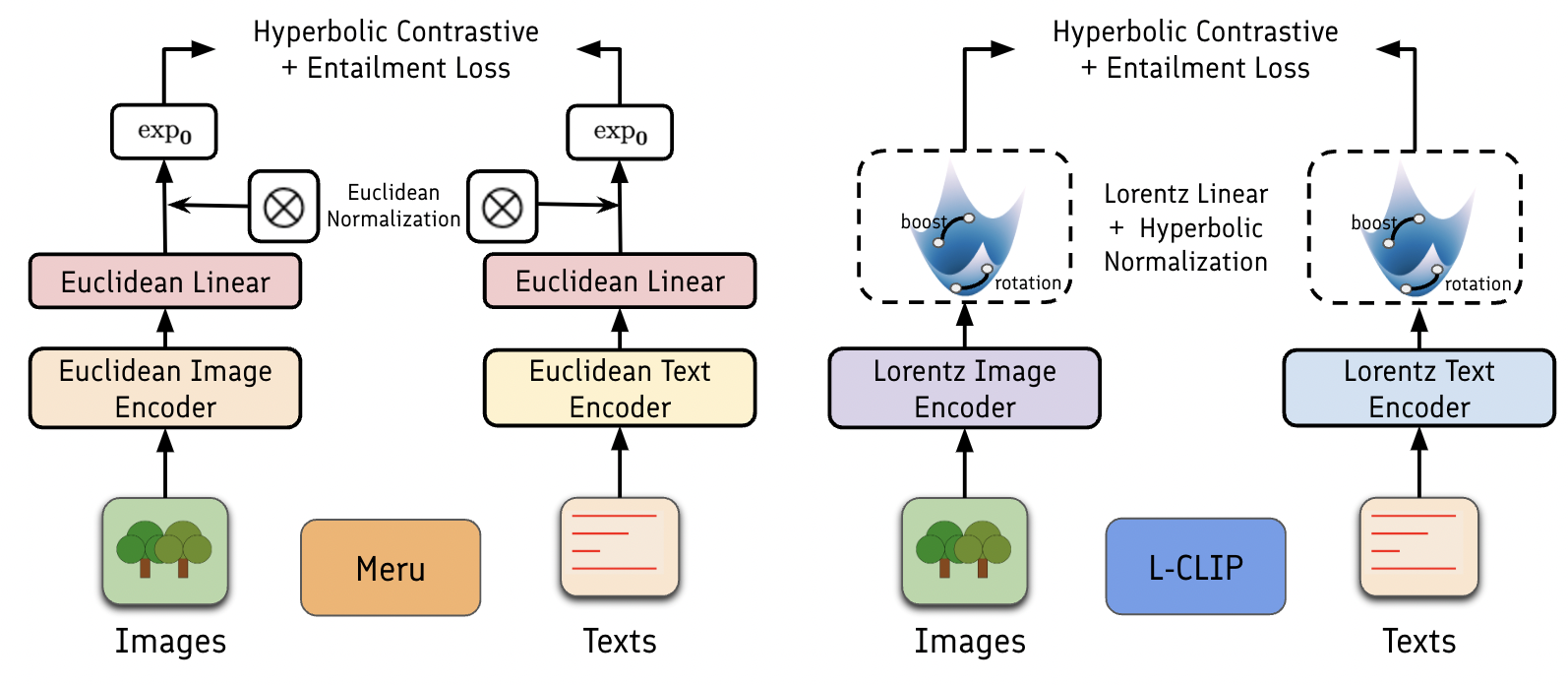}
       \caption{Framework of Lorentzian CLIP model (L-CLIP) for multi-modality learning (right), consisting of a hyperbolic image and text encoder. We build the first hyperbolic CLIP model using \name with LViT as the image encoder and a hyperbolic language Transformer as the text encoder. In comparison, MERU (left) is a hybrid model, utilizing Euclidean encoders and  computes the loss in hyperbolic space.}
       \label{fig:LCLIP}
\end{figure}

\section{Building New Hyperbolic Foundation Models}\label{hvit}

\name's comprehensive components can be intuitively put together with minimal efforts to develop new hyperbolic foundation models and fine-tuning pipelines. In this section, we demonstrate the ease of creating three hyperbolic counterparts of Euclidean foundation models by detailing the construction of two novel fully hyperbolic foundation models in Lorentz space not seen in literature: (1) a \textit{fully hyperbolic vision Transformer (LViT)}, which is then fine-tuned with hyperbolic LoRA, using our \name framework; (2) a \textit{fully hyperbolic CLIP model (L-CLIP)} which enables hyperbolic multi-modal learning; and (3) a \textit{hyperbolic graph retrieval-augmented generation pipeline (HypGraphRAG)}~\cite{edge2024graphrag}, demonstrating \name's  support for graph foundation models. How these models fit into the taxonomy of models supported by \name and a subset of the layers used are shown in \cref{model_graph}.

To the best of our knowledge, fully hyperbolic variants of vision Transformers and CLIP models have not been studied or proposed in prior works. In the context of hyperbolic vision Transformers, HVT~\cite{fein-ashley2024hvt} is not a fully hyperbolic ViT, as many of its operations still rely on the tangent space (e.g., LayerNorm, residual connections, etc.).  For CLIP models, MERU~\cite{desai2023hyperbolic} is a hybrid approach that computes the loss in hyperbolic space while retaining Euclidean image and text encoders. This is also the first work that builds a hyperbolic GraphRAG model. The performance of LViT in image classification tasks, with and without fine-tuning, is evaluated in \cref{img_class}. We also evaluate the performance of L-CLIP for zero-shot image and text retrieval tasks in \cref{exp:retrieval} and the performance of HypGraphRAG for question-answering tasks in \cref{exp:graph_rag}.

\textbf{Lorentz ViT Model Architecture.} The overall model architecture is shown in \cref{fig:HViT}. Before building the hyperbolic ViT model, the input images are first projected into the Lorentz space via an exponential map at the tangent space of the origin. The embedded image data is then transformed through a hyperbolic patch embedding layer, built on top of Lorentz convolutional~\cite{Bdeir2024fully}. The patch embeddings are then combined with a (possibly learned) hyperbolic positional encoding, where the addition is done with LResNet~\cite{he2025lresnet}. The transformed patch embeddings are then passed through a Lorentz ViT encoder, where each layer consists of a Lorentz self-attention layer~\cite{chen2021fully} followed by a 2-layer Lorentz MLP. The encoder output is then average pooled across the patches~\cite{Bdeir2024fully} before passing through a classifier, which is a Lorentz MLR layer~\cite{Bdeir2024fully} for classification. Lorentz layer normalization~\cite{yang2024hypformer}, dropout~\cite{yang2024hypformer}, activation~\cite{yang2024hypformer}, and residual connection~\cite{he2025lresnet} are implemented in appropriate places similar to the Euclidean ViT~\cite{dosovitskiy2020image}. 

\textbf{Lorentz CLIP Model Architecture.} Similar to Euclidean CLIP models~\cite{radford2021CLIP}, the hyperbolic CLIP model consists of an image encoder and a text encoder, as shown in \cref{fig:LCLIP}. Compared to MERU~\cite{desai2023hyperbolic}, which is a hybrid CLIP model where only the loss being computed in hyperbolic space but still uses Euclidean encoders, \name enables fully hyperbolic CLIP models with its support for hyperbolic foundation models in both text and image modalities. As a demonstration of \name's capabilities, we build the fully hyperbolic first multi-modal foundation model with a hyperbolic vision Transformer and text Transformer as the image and text encoders respectively. The former is the LViT model we detailed above. The later is a fully hyperbolic language Transformer, comprising of hyperbolic word and positional encodings, multi-head attention layers, feed forward MLP layers, and other essential modules such as layer normalization. Hyperbolic Transformers in prior works lacked some of these components~\cite{chen2021fully, HNN++, gulcehre2019hyperbolicAT} when compared to the one we build here. For instance, none of them implemented a hyperbolic layer normalization layer.

\begin{figure} 
    \centering
        \centering
\includegraphics[width=0.85\linewidth]{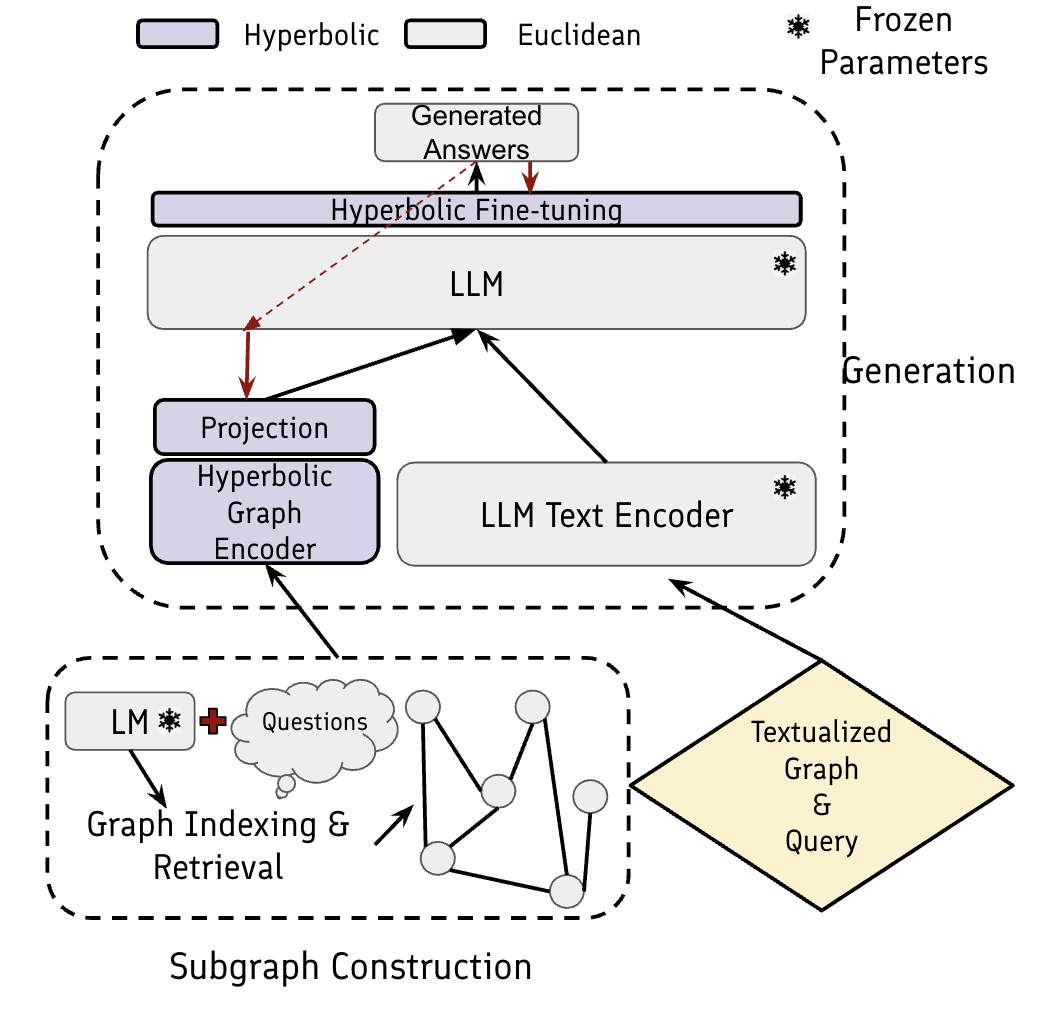}
       \caption{Framework of hyperbolic GraphRAG model (HypGraphRAG). Compare to standard Euclidean GraphRAG, a hyperbolic graph encoder is employed to encode the retrieved subgraph and (optional) hyperbolic LoRA for fine-tuning.}
       \label{fig:HypGraphRAG}
\end{figure}

\textbf{Hyperbolic Graph RAG.} \name can be used to build a wide range of other hyperbolic foundation models. For instance, \name's capability to build popular hyperbolic LLMs and hyperbolic GNNs can be put together to build the first hyperbolic Graph RAG~\cite{edge2024graphrag} pipeline for retrieval and generation. \cref{fig:HypGraphRAG} shows HypGraphRAG that extends the Euclidean model to incorporate a hyperbolic GNN and hyperbolic fine-tuning. We also demonstrate the comprehensive support for hyperbolic GNNs in \name with an extensive benchmark in \cref{bench}. 

\textbf{Simplifying and Reducing Redundancy in Future Research.} With its comprehensive modules that enables effortless construction of new hyperbolic foundation models, \name eliminates the need to search previous literature for relevant hyperbolic operations or reconcile incompatible methods. For instance, many components of the recently proposed Poincar{'e} Vision Transformer (HVT)\cite{fein-ashley2024hvt} were originally introduced in prior studies. As a result, HVT can be mostly constructed from \name's modules. In cases like these, \name enables future research to instead focus more on analyzing hyperbolic models for insights rather than spending excessive resources and energy to construct the model. Additionally, \name would reduce the likelihood of redundant proposals for similar hyperbolic operations and models.

 \begin{table}

\caption{Test accuracy (ACC) ($\%$) for Image Classification task on CIFAR-10 and CIFAR-100  with hyperbolic ResNets. We also show ACC for hyperbolic vision Transformers on the CIFAR and ImageNet-1K datasets. The $\delta$ values are taken from HCNN~\cite{Bdeir2024fully}, which are normalized by the diameter of the dataset.}
\label{Classification Results}
\centering
\resizebox{0.47\textwidth}{!}{%
\begin{tabular}{@{}lccccc@{}}
\toprule
\textbf{Dataset}                            & \sc{CIFAR-10} & \sc{CIFAR-100} &\sc{Tiny-ImageNet} & \sc{ImageNet} \\ 
\textbf{Hyperbolicity}                     &$\delta=0.26$ &$\delta=0.23$&$\delta=0.20$ & -\\
\midrule
HCNN~\cite{van2023poincar}         & $95.02\pm0.19$             & $77.31\pm 0.21$  & $65.01\pm 0.29$      & -    \\ 
Poincar{\'e} ResNet~\cite{hgcn2019}              & $94.71\pm0.13$             & $76.91\pm 0.34$ & $63.11\pm 0.59$     & -       \\
\midrule
ViT~\cite{dosovitskiy2020image} & 98.13 &87.13 & - & 77.91 \\
HVT~\cite{fein-ashley2024hvt}     &    $61.44$         &     $42.77$ & $40.12$  & $78.2$       \\
LViT (built by us)      &     $85.02$   & $69.11$ & $53.01$ & $\mathbf{79.4}$ \\

LViT (fine-tuned w/ HypLoRA) &   $\mathbf{98.18}$   &  $\mathbf{87.36}$  &  $\mathbf{74.11}$ & $\mathbf{79.4}$\\
 \bottomrule
\end{tabular}%
}
\end{table}

\section{Empirical Evaluations}\label{eval}
In this section, we evaluate the performance of the LViT and L-CLIP models built in \cref{hvit}. Additionally, to evaluate the correctness of our implementation and demonstrate its comprehensiveness, we perform several empirical experiments across graph, image, and language modalities. These experiments include both evaluation of the new hyperbolic models built in \cref{hvit} and recreation of prior works to demonstrate scope. The corresponding model-task combinations in \cref{model_graph} illustrate how the foundation model, the evaluated downstream task, and the associated modality integrate into the overall structure of \name and its modules.
For experiments that recreate prior works, we try to follow the original implementations and hyperparameter setups of the respective paper as closely as possible to reproduce their results, with one except being the hyperbolic GNN benchmark discussed in \cref{bench}. The purpose of our empirical experiments is not to achieve SOTA on any particular method, but rather to demonstrate \name's versatility and comprehensiveness across foundation model architectures and modalities. Additional details are available at \cref{appendix:exp}.

\subsection{Hyperbolic ViT and CNNs}\label{img_class}
In this section, we demonstrate \name's comprehensive support for the image modality by evaluating the performance of hyperbolic CNNs and ViTs.

\textbf{Evaluating LViT.} We evaluate the image classification accuracy of the Lorentz ViT we built in \cref{hvit} on 4 datasets: CIFAR-10\cite{Krizhevsky2009Learning}, CIFAR-100\cite{Krizhevsky2009Learning}, Tiny-ImageNet~\cite{tinyimagenet}, and ImageNet-1K~\cite{deng2009imagenet}. We compare its performance with HVT~\cite{fein-ashley2024hvt}, implemented using \name with the original experimental setup. We also fine-tune LViT on the smaller datasets  using HypLoRA~\cite{yang2024hyplora}, where the model was pre-trained on the ImageNet dataset, and compare with the fine tuned results of the Euclidean ViT~\cite{dosovitskiy2020image}. 

\textbf{Reproducing Hyperbolic CNNs.} We further reproduce recently developed hyperbolic CNNs using the ResNet-18 architecture \cite{he2016deep}, staying true to the model and training setup in prior works~\cite{van2023poincar, Bdeir2024fully}. While we focus on reproducing the results from previous studies, \name is \textit{not} limited to these ResNet models. Additional ResNet models in the Lorentz space can be built using the improved Lorentzian linear layer that allows for per-layer change of curvature~\cite{yang2024hypformer} and more effective residual connections~\cite{he2025lresnet}. 

The results are shown in \cref{Classification Results}, where we report the classification accuracy (\%). The results are largely similar to the ones in HCNN~\cite{Bdeir2024fully} for the hyperbolic CNNs. On the smaller datasets, the hyperbolic ViTs face difficulties due to the inherent lack of inductive bias. The fully hyperbolic LViT outperforms HVT, which contains tangent space based operations. On the ImageNet dataset, both hyperbolic ViTs perform well and better than the Euclidean ViT. With fine-tuning, the pre-trained LViT is the best performer for all 4 datasets. 
\begin{table}
\caption{BLEU score for previous hyperbolic Transformers on machine translation (MT) tasks on IWSLT’14 and WMT’14, in low dimensional settings with embedding dimensions 64 and 128 for both datasets.}
\label{Transformer Results}
\centering
\resizebox{0.40\textwidth}{!}{%
\begin{tabular}{lcccc}
\toprule
\textbf{Dataset}& \multicolumn{2}{c}{\sc{IWSLT’14}} & \multicolumn{2}{c}{\sc{WMT’14}}  \\ 
\textbf{Dim}                     &$D=64$ &$D=128$ & $D=64$ &$D=128$\\
\midrule
HNN++\cite{HNN++} &$23.1$ & $24.9$ & $18.0$  &    $21.6$         \\
HAN\cite{gulcehre2019hyperbolicAT} &$23.5$ & $25.5$ & $18.9$  &   $22.0$          \\
HyboNet\cite{chen2021fully} &$\mathbf{25.2}$ &$\mathbf{26.1}$  & $\mathbf{19.1}$  &   $\mathbf{22.7}$          \\
 \bottomrule
\end{tabular}%
}
\end{table}

\begin{table}
\caption{Accuracy of Gemma-7B and LLaMA3-8B fine tuned with HypLoRA on arithmetic reasoning tasks. The percentage following each dataset indicates the proportion of prompts relative to the total number of inference prompts.}
\label{tab:hypLoRA}
\resizebox{0.44\textwidth}{!}{%
\begin{tabular}{@{}lccc@{}}
\toprule
\textbf{Model} & \sc{MAWPS($8.5\%$)} & \sc{GSM8K($46.9\%$)} & \sc{AQuA($9.0\%$)} \\ \midrule
Gemma-7B~\cite{gemma2024}       & $91.2$         & $68.7$          & $32.9$        \\
LLaMA3-8B~\cite{dubey2024llama3}      & $\mathbf{91.5}$         & $\mathbf{73.3}$          & $\mathbf{34.3}$        \\ \bottomrule
\end{tabular}
}
\end{table}

\begin{table*}[t]
\caption{Test ROC AUC results (\%) for Link Prediction (LP),  F1 scores (\%) for Node Classification (NC) for homophilous graph, and mAP (\%) for MD tasks. A lower 
$\delta$ value indicates a more tree-like dataset and is sourced from HGCN~\cite{hgcn2019}. For MD tasks, standard deviations are within 0.1 and are therefore omitted. Best performances are bolded}
\label{graph table}
\centering
\resizebox{0.90\textwidth}{!}{%
\begin{tabular}{lcccccccccc}
\toprule
                   \textbf{Dataset}& \multicolumn{2}{c}{\sc{Disease}} & \multicolumn{2}{c}{\sc{Airport}} & \multicolumn{2}{c}{\sc{PubMed}} & \multicolumn{2}{c}{\sc{Cora}} & \sc{Power}      & \sc{Bio-Worm} 
                   \\
                   \textbf{Hyperbolicity}& \multicolumn{2}{c}{$\delta=0$} & \multicolumn{2}{c}{$\delta=1$} & \multicolumn{2}{c}{$\delta=3.5$} & \multicolumn{2}{c}{$\delta=11$} & \sc{-}      & \sc{-}\\ 
\textbf{Task}      & LP                             & NC                               & LP                              & NC                              & LP                               & NC                              & LP                             & NC                             & MD                       & MD                       \\ \midrule
HNN~\cite{HNN}           
& ${94.3 \pm 0.1}$                
& ${37.1 \pm 2.7}$                  
& ${93.4 \pm 0.1}$          
& ${88.5 \pm 0.1}$                 & ${93.2 \pm 0.1}$                  & ${72.7 \pm 0.7}$                 & ${91.3 \pm 0.1}$                & ${51.2 \pm 1.2}$                & ${82.2}$
& ${78.6}$  
\\
HGCN~\cite{hgcn2019} & ${93.6 \pm 0.1}$                & ${90.8 \pm 1.2}$                  & ${96.5 \pm 0.2}$                 & ${91.1 \pm 1.4}$                 & ${94.4 \pm 0.1}$         & ${71.1 \pm 0.3}$                 & ${93.5 \pm 0.1}$       & ${80.4 \pm 0.5}$                  & ${90.1}$          & ${78.8}$          \\
LGCN~\cite{lgcn}      & ${97.8 \pm 0.5}$                  & ${90.3 \pm 0.9}$                  & ${96.6 \pm 0.2}$                 & ${90.1 \pm 0.8}$                   & ${95.5 \pm 0.2}$                  & ${78.1 \pm 0.5}$                 & ${93.5 \pm 0.4}$                & ${78.8 \pm 1.2}$                & ${94.8}$          & ${87.6}$          \\
HyboNet~\cite{chen2021fully}     & ${97.7 \pm 0.5}$                & ${\textbf{94.1} \pm 1.0}$                  & ${96.6 \pm 0.2}$                 & ${\textbf{93.1} \pm 0.9}$                 & ${96.3 \pm 0.1}$                  & ${79.0 \pm 1.3}$                 & ${93.9 \pm 0.4}$                & ${\textbf{80.1} \pm 1.5}$                & ${96.0}$          & ${88.0}$ 
\\
HGNN~\cite{liu2019HGNN}    & ${71.7 \pm 1.7}$          & ${79.3 \pm 1.5}$         & ${\textbf{97.5} \pm 0.1}$        & ${90.9 \pm 1.0}$        & ${93.8 \pm 0.1}$                  & ${78.6 \pm 0.9}$        & ${91.9 \pm 0.2}$                & ${\textbf{79.8} \pm 0.6}$       & ${\textbf{96.7}}$ & ${92.0}$ \\ 
HGAT~\cite{gulcehre2019hyperbolicAT}    & ${85.3\pm 0.2}$          & ${88.3 \pm 0.9}$         & ${94.0 \pm 1.0}$        & ${89.8 \pm 0.7}$        & ${93.1 \pm 0.3}$                  & ${78.0 \pm 1.3}$        & ${93.9 \pm 0.2}$                & ${78.5 \pm 1.4}$       & ${96.0}$ & ${\textbf{94.7}}$ \\
GIL~\cite{zhu2020gil}    & ${\textbf{99.9} \pm 0.2}$          & ${90.0 \pm 0.7}$         & ${97.3 \pm 0.3}$        & ${91.2 \pm 1.1}$        & ${\textbf{97.3} \pm 0.9}$                  & ${83.2 \pm 0.9}$        & ${\textbf{95.8} \pm 0.2}$                & ${79.3 \pm 1.1}$       & ${96.1}$ & ${90.0}$ \\
\bottomrule
\end{tabular}%
}
\end{table*}

\subsection{Hyperbolic Transformers}
Using \name, we replicate previous hyperbolic transformer experiments on the natural language modality, demonstrating \name's support for diverse modalities. 
We implement the hyperbolic Transformer model from HyboNet~\cite{chen2021fully}, HAN~\cite{gulcehre2019hyperbolicAT}, and HNN++~\cite{HNN++} using \name to reproduce their experiment on the IWSLT’14 English-German and WMT’14 English-German benchmarks on low dimensional settings, with embedding dimensions of 64 and 128. We report the BLEU score. Here the fully hyperbolic Transformer architecture (HyboNet) outperforms the tangent space-dependent and partially hyperbolic ones (HAN and HNN++).

\subsection{Hyperbolic Fine Tuning of LLMs}
To demonstrate \name's support for building hyperbolic tuning pipelines for LLMs, we also reproduce results from HypLoRA~\cite{yang2024hyplora}, where we fine-tune both Gemma-7B~\cite{gemma2024} and LLaMA3-8B~\cite{dubey2024llama3} with rank $r = 32$. The Euclidean inputs are normalized before lifting to the manifold with an exponential map at the origin as did in previous studies~\cite{yang2024hyplora, desai2023hyperbolic}. As does in the original HypLoRA study, the training set consists of data from GSM8K~\cite{cobbe2021gsm8k}, MAWPS, MAWPS-single~\cite{koncel2016mawps}, 1,000 examples from AQuA~\cite{goswami2023aqua}, and the math-10K dataset consisting of step-by-step rationales generated by ChatGPT~\cite{yang2024hyplora}. The testing set for evaluation consists of the training set consists of GSM8K~\cite{cobbe2021gsm8k}, MAWPS~\cite{koncel2016mawps}, and AQuA~\cite{goswami2023aqua}. While the same datasets were used for training and testing, there are no overlapping data points. We closely follow the original implementation and use the same hyperparameters. We also remove the additional time-like dimension from the final output, which is the leading $0$ for vectors on the tangent space of the origin, as did in the original study. The results are shown in \cref{tab:hypLoRA}. We find that the results are similar to the original paper.  

\subsection{Benchmarking Hyperbolic GNNs}\label{bench}
We evaluate the performance of hyperbolic GNNs on node classification (NC), link prediction (LP), and graph reconstruction tasks (MD). The graph reconstruction task was not considered by most previous works on hyperbolic GNNs and were added as an additional benchmark for our testing. We further consider the performance of HNN~\cite{HNN} as well. We utilize several datasets: (1) For link prediction and node classification tasks, we utilize four common datasets in prior works of hyperbolic GNNs, which are {\sc Disease}\cite{hgcn2019}, {\sc Airport}\cite{hgcn2019}, {\sc PubMed}\cite{sen2008collective}, and {\sc Cora}\cite{sen2008collective}; (2) For MD tasks, we test on two real-world datasets: {\sc Power}~\cite{watts1998nature} and {\sc Bio-Worm}~\cite{cho2014wormnet}. For link prediction and node classification tasks, we do a search for hyperparameters, particularly for curvature ranging from $-0.25$ to $-2.0$, and report the F1-score for the NC task and the Area Under Curve (AUC) for the LP task. For graph reconstruction, we use a constant curvature of $-1.0$ and report the mean average precision (mAP) value.

The results are shown in \cref{graph table}. For link prediction and node classification, the results were generally comparable and/or better than the ones in the respective papers. In some cases, such as link prediction on the {\sc disease} dataset for HNN, the performance benefited significantly from the hyperparameter search on curvatures compared to the original paper~\cite{HNN}. The graph reconstruction tasks were not considered by most existing works on hyperbolic GNNs, meaning that there are no reference performances to compare with. Interestingly, the performance for link prediction and node classification tasks are not necessarily good indicators for the model performance on graph reconstruction. However, this could be due to the lack of a comprehensive hyperparameter search for this task. 

\subsection{Constrastive Pre-training with L-CLIP}\label{exp:retrieval}
To demonstrate \name's capability in building hyperbolic multi-modal foundation models, we test the hyperbolic CLIP (L-CLIP) model we build in \cref{hvit} by pre-training the model using image-text pairings and transferring to image-text retrieval tasks. We largely follow the training scheme in MERU~\cite{desai2023hyperbolic}, with the combination of hyperbolic contrastive and entailment loss. As training on the 14-million images in the full RedCaps~\cite{desai2021redcaps} dataset used by MERU requires extensive computational resources and time, we randomly sampled 10\% of the RedCaps dataset and pre-train L-CLIP on this subset as a demonstration of concept, where we used the pre-trained LViT model from \cref{img_class} as our image encoder. We evaluate the trained L-CLIP model on zero-shot image and text retrieval tasks on the COCO benchmark~\cite{chen2015cococaptions}, where the transferring is done without any additional training. The recall@\{5,10\} results for text-to-image retrieval are $28.0$ and $38.1$ respectively. 
While the goal here is not to demonstrate SOTA performance, the successful training of L-CLIP demonstrates its support for pre-training hyperbolic multi-modal foundation models. 

\subsection{Hyperbolic Graph RAG}\label{exp:graph_rag}
To further demonstrate \name's support for building multi-modal and graph foundation models, we test its capability in building graph retriever models. Here we test the hybrid Graph RAG model~\cite{he2024gretriever}, the HypGraphRAG model proposed in \cref{hvit}, where the graph encoder is implemented with a skip-connected GNN~\cite{zhang2021hyperbolic} and the LLM is fine-tuned with HypLoRA~\cite{yang2024hyplora}. Following the experiments in the Graph RAG paper, we test the Hit@1 accuracy for question-answering tasks in a graph QA dataset, namely the WebQSP~\cite{yih2016value} datasets. We used LLaMa3.1-8B~\cite{dubey2024llama3} as our LLM. Our hybrid Graph RAG achieved $73.89\pm1.09$ Hit@1 accuracy for the WebQSP dataset, again demonstrating \name's comprehensive scope and support for foundational models.

\section{Conclusion}
We introduce \name, a comprehensive framework comprised of the core modules for building hyperbolic foundation models. Built on top of PyTorch~\cite{paszke2019pytorch} with a similar design philosophy, \name provides accessible modules that make it simple and intuitive to build hyperbolic foundation models for researchers and the general AI audience alike. Compared to existing frameworks such as HypLL~\cite{spengler2023hypll} and Hyperlib~\cite{hyperlib}, \name emphasizes its comprehensiveness and support for building novel hyperbolic foundation models for diverse modalities. As a demonstration, we proposed and tested the first hyperbolic vision Transformer with fine-tuning pipelines, the first hyperbolic CLIP model, and extended GraphRAG to incorporate hyperbolic graph encoder and fine-tuning. Our experiments demonstrated the extensive support \name has for building new hyperbolic foundation models, enabling future research to instead focus on insightful analysis.
\newpage
\bibliographystyle{ACM-Reference-Format}
\bibliography{main}
\appendix
\section{Preliminary: Hyperbolic Geometry}
This section provides an overview of the fundamental concepts in hyperbolic geometry, focusing on the Lorentz model and Poincar{\'e} ball model

\subsection{Lorentz Model} 
An $n$-dimensional Lorentz model is
a Riemannian manifold $(\mathcal{L}^n, \mathfrak{g}_n^K)$ equipped with the Riemannian metric tensor $\mathfrak{g}_n^K = \mathrm{diag}(-1, 1, \ldots, 1)$ and defined by a constant negative curvature $K<0$, denoted as $\mathbb{L}^{K,n}$. Each point $\x\in\mathbb{L}^{K,n}$ has the parametrized form $[x_t, \x_s]^T$ where $x_t\in\R$ is called the time-like component and $\x_s\in\R^{n}$ is called the space-like component. $\mathbb{L}^{K,n}$ is equipped with the \textit{Lorentzian inner product}. For points $\x,\y\in\mathbb{L}^{K,n}$, their inner product $\langle\x,\y\rangle_\mathcal{L}$ is given by 
\begin{align}
    \langle\x,\y\rangle_{\mathcal{L}} &= -x_ty_t + \x_s^T\y_s = \x^T\mathfrak{g}_n^K\y,
\end{align}
with $|\|\x\||_\mathcal{L}\coloneq\sqrt{|\langle \x, \x\rangle_\mathcal{L}|}$ being the Lorentzian norm. Formally, $\mathcal{L}^n$ is the point set \begin{displaymath}
    \mathcal{L}^n = \{\x\in\R^{n+1}: \langle\x,\x\rangle_\mathcal{L} = 1/K, x_t>0\}.
\end{displaymath}
The origin $\mathbf{o}\in\mathbb{L}^{K,n}$ is the point $[\sqrt{-1/K}, 0,\ldots,0]^T$.

\textbf{Tangent space.}
The tangent space at a point $\x\in\mathbb{L}^{K,n}$ is set of points orthogonal to $\x$, defined as \begin{displaymath}
    \mathcal{T}_\mathbf{x}\mathbb{L}^{K,n} = \{\y\in\R^{n+1}: \langle\x,\y\rangle_{\mathcal{L}} =0 \}.
\end{displaymath}
Notably, the tangent space is isometric to Euclidean space.

\textbf{Exponential and logarithmic maps. }
For each point $\x\in\mathbb{L}^{K,n}$, the exponential map $\exp_\x^K:\mathcal{T}_\x\mathbb{L}^{K,n}\to \mathbb{L}^{K,n}$ and the logarithmic map $\log_\x^K:\mathbb{L}^{K,n}\to \mathcal{T}_\mathbf{x}\mathbb{L}^{K,n}$ at $\mathbf{x}$ are given by
\begin{equation}
        \exp_\x^K(\y) = \cosh(\alpha)\x + \frac{\sinh(\alpha)}{\alpha}\y,
 \alpha = \sqrt{-K\langle\x,\y\rangle_\mathcal{L}},
\end{equation}
\begin{equation}
   \log_\x^K(\x)= \frac{\cosh^{-1}(\beta)}{\sqrt{\beta^2-1}}(\y-\beta\x),
     \beta = K\langle\x,\y\rangle_\mathcal{L}.
\end{equation}

\textbf{Parallel transport.}
    Parallel transport is a generalization of translation to hyperbolic geometry mapping a point $\z\in\mathcal{T}_\x\mathbb{L}^{K,n}$ to a point in $\mathcal{T}_\y\mathbb{L}^{K,n}$ via \begin{equation*}
        \mathbf{P}_{\x\to\y}(\z) = \z + \frac{\langle\y,\z\rangle_\mathcal{L}}{-1/K-\langle\x,\y\rangle_\mathcal{L}}(\x+\y).
\end{equation*}

$\mathbf{P}_{\x\to\y}(\z): \mathcal{T}_\x\mathbb{L}^{K,n}\to \mathcal{T}_\y\mathbb{L}^{K,n}$

\subsection{Poincar{\'e} Ball Model}
The Poincar{\'e} ball model of hyperbolic space, $\mathbb{P}^{n, K}$ is the $n$-dimensional sphere $S^n$ with radius $1/\sqrt{K}$, with the Riemannian metric $g_x^{\mathbb{P}} = \lambda_x^2 g^E$, where $ \lambda_x := \frac{2}{1 - c\|x\|^2}$ and $\mathfrak{g}^E$ is the Euclidean metric. The induced distance is given by \[
d_c(x,y) = \left( \frac{2}{\sqrt{c}} \right) \tanh^{-1} \left( \sqrt{c} \| -x \oplus_c y \| \right).
\]
The exponential map and logarithmic map on the ball are given by 
\[
\exp_x^c(v) = x \oplus_c \left( \tanh \left( \frac{\sqrt{c} \lambda_x^c \|v\|}{2} \right) \frac{v}{\sqrt{c} \|v\|} \right),
\]
and 
\[
\log_x^c(y) = \frac{2}{\sqrt{c} \lambda_x^c} \tanh^{-1} \left( \sqrt{c} \| - x \oplus_c y \| \right) \frac{- x \oplus_c y}{\| - x \oplus_c y \|}.
\]
respectively, where $\oplus_c$ denotes the mobius addition given by
\[
x \oplus_c y := \frac{(1 + 2c \langle x, y \rangle + c \|y\|^2)x + (1 - c \|x\|^2)y}{1 + 2c \langle x, y \rangle + c^2 \|x\|^2 \|y\|^2}.
\]
The parallel transport map is given by 
\[
P_{x \to y}^{c}(v) = \frac{\lambda_x^c}{\lambda_y^c} \operatorname{gyr}[y, -x] v.
\]

\section{Experiment Details}\label{appendix:exp}
In this section we provide additional details on the experiments in \cref{eval}, focusing on the new hyperbolic foundation models we build in this study. All experiments were implemented on a combination of RTX 2080Ti and V100 GPUs. 
\subsection{Lorentz Vision Transformer}
We model our construction of LViT after the base ViT model from~\cite{dosovitskiy2020image}. To this end, LViT is built with $12$ heads and an embedding dimension of $768$, resulting in $64$ dimensions per head. The feed-forward MLP consists of $2$ layers with a hidden dimension of $3072$. Note that as the Lorentz space is an $n$-dimensional manifold embedded in an $(n + 1)$-dimensional ambient space, a $k$-dimensional embedding here refers to the $k$-dimensional Lorentz vector embedded in the $(k+1)$-dimensional ambient space. For our experiment on ImageNet, the images are cropped to be of size $224\times 224$, and a patch size of $16$ was used, just like the Euclidean case. For the CIFAR datasets, the images are of size $32\times 32$ with a patch size of $4$. For TinyImageNet, the images are of size $64$ with a patch size of $8$. For all experiments, we used the AdamW optimizer for the Euclidean parameters and the Riemannian Adam optimizer for the manifold parameters. The relevant hyperparameters were obtained through a grid search over the following choices: \textit{Euclidean learning rate} in \{1e-4, 3e-4, 1e-3, 5e-3\}, \textit{hyperbolic learning rate} in \{1e-4, 3e-4, 1e-3, 5e-3\}, \textit{Euclidean weight decay} in \{0, 1e-3, 1e-2\}, \textit{hyperbolic weight decay} in \{0, 1e-4, 1e-3\}, \textit{MLP dropout} in \{0, 0.1\}, and \textit{attention dropout} in \{0, 0.1\}. Additionally, in our residual connection enabled by LResNet~\cite{he2025lresnet}, we used learnable weights with a high scaling factor of 25-30. We fixed the curvature of all manifolds to be $-1$. 

\subsection{Lorentz CLIP Model}
Here we expand on the details regarding the hyperbolic CLIP model L-CLIP. The visual encoder we used was the LViT model pre-trained on ImageNet from the above section. For the text encoder, we built a hyperbolic language Transformer encoder following the setup in MERU~\cite{desai2023hyperbolic}. Specifically, we built a 12-layer, 512 dimensions wide Transformer language model. We use the same byte-pair encoding tokenizer as did in MERU and CLIP, and truncate input text at maximum 77 tokens. We fixed the curvature of the language Transformer at $-1$ as well. The hyperparameters used were large the same as MERU and we followed the same training setup as well. 

\subsection{Lorentz Graph RAG}
For the hybrid Graph RAG model, the GNN graph retriever was implemented using the same skip-connected GNN architecture from LResNet~\cite{he2025lresnet}, where we used $4$-layers of the fully hyperbolic graph convolutional layer from HyboNet~\cite{chen2021fully} with the residual connection. The curvature of the graph retriever was fixed at $-1$ as well. The LLM, which is an LLaMA3.1-8B model, was fine-tuned with HypLoRA using a learnable curvature, a rank of $8$, and an alpha of $16$. The hyperparameters and training setup largely followed the Euclidean Graph RAG~\cite{he2024gretriever} and the PyTorch Geometric implementation. 

\subsection{GNN Benchmark}
Here we provide more details about our GNN benchmark experiment. For the link prediction and node classification tasks, we performed a comprehensive search over \textit{learning rate, weight decay, and the dropout rate} with a fixed embedding dimension of $16$. Additionally, we performed a comprehensive search over the curvature of the model over the set $\{-2,0, -1.5, -1.0, -0.5, -0.25, \text{trainable}\}$. The training setup largely followed the one from HGCN~\cite{hgcn2019}. 

For the graph reconstruction task, we followed the setup and hyperparameters in QGCN~\cite{xiong2021pseudo}, with an embedding dimension of $10$, $2$ hidden layers, learning rate of $0.01$, and learnable curvature.

When training models with learnable curvatures, we allowed for per-layer varying curvature whenever the underlying model architecture allowed. 

\end{document}